\documentclass[10pt,twocolumn,letterpaper]{article}

\usepackage[pagenumbers]{cvpr} % To force page numbers, e.g. for an arXiv version

\usepackage{graphicx}
\usepackage{amsmath}
\usepackage{amssymb}
\usepackage{booktabs}
\usepackage{adjustbox}
\usepackage{pifont}

\newcommand{\cmark}{\ding{51}}
\newcommand{\xmark}{\ding{55}}

\usepackage[accsupp]{axessibility}  % Improves PDF readability for those with disabilities.

\usepackage[dvipsnames]{xcolor}
\usepackage{xcolor}
\usepackage{colortbl} % for tab. color
\def\rowg{\rowcolor{gray!10}}
\def\cellg{\cellcolor{gray!10}}

\def\supp{\textit{\textcolor{BrickRed}{supplementary materials}}}

\usepackage[pagebackref,breaklinks,colorlinks]{hyperref}

\usepackage[capitalize]{cleveref}
\crefname{section}{Sec.}{Secs.}
\Crefname{section}{Section}{Sections}
\Crefname{table}{Table}{Tables}
\crefname{table}{Tab.}{Tabs.}

\def\cvprPaperID{4308} % *** Enter the CVPR Paper ID here
\def\confName{CVPR}
\def\confYear{2023}

\begin{document}
\title{Category Query Learning for Human-Object Interaction Classification}

\author{
Chi~Xie$^1$\footnotemark[2] \quad
Fangao Zeng$^{2}$\footnotemark[3] \quad
Yue~Hu$^3$\footnotemark[3] \quad
Shuang~Liang$^{1}$\footnotemark[1] \quad
Yichen~Wei$^{2}$\footnotemark[3] \\
$^{1}${Tongji University} \quad
$^{2}${MEGVII Technology} \quad
$^{3}${Shanghai Jiao Tong University} \\
$^{1}${\tt\small \{chixie, shuangliang\}@tongji.edu.cn} \\ \quad
$^{2}${\tt\small zfg472988436@163.com, wei\_yi\_chen@hotmail.com} \quad
$^{3}${\tt\small 18671129361@sjtu.edu.cn} \\
}

\maketitle

\renewcommand{\thefootnote}{\fnsymbol{footnote}}
\footnotetext[1]{Corresponding author.}
\footnotetext[2]{Work done during internship at MEGVII technology.}
\footnotetext[3]{Work done while worked at MEGVII.}

%%%%%%%%% ABSTRACT
\begin{abstract}
Unlike most previous HOI methods that focus on learning better human-object features, we propose a novel and complementary approach called \emph{category query learning}. Such queries are explicitly associated to interaction categories, converted to image specific category representation via a transformer decoder, and learnt via an auxiliary image-level classification task. This idea is motivated by an earlier multi-label image classification method, but is for the first time applied for the challenging human-object interaction classification task. Our method is simple, general and effective. It is validated on three representative HOI baselines and achieves new state-of-the-art results on two benchmarks. Code will be available at \url{https://github.com/charles-xie/CQL}.
\end{abstract}

%%%%%%%%% BODY TEXT
\section{Introduction}
\label{sec:intro}
\begin{figure}[t]
    \centering
    \includegraphics[width=\linewidth]{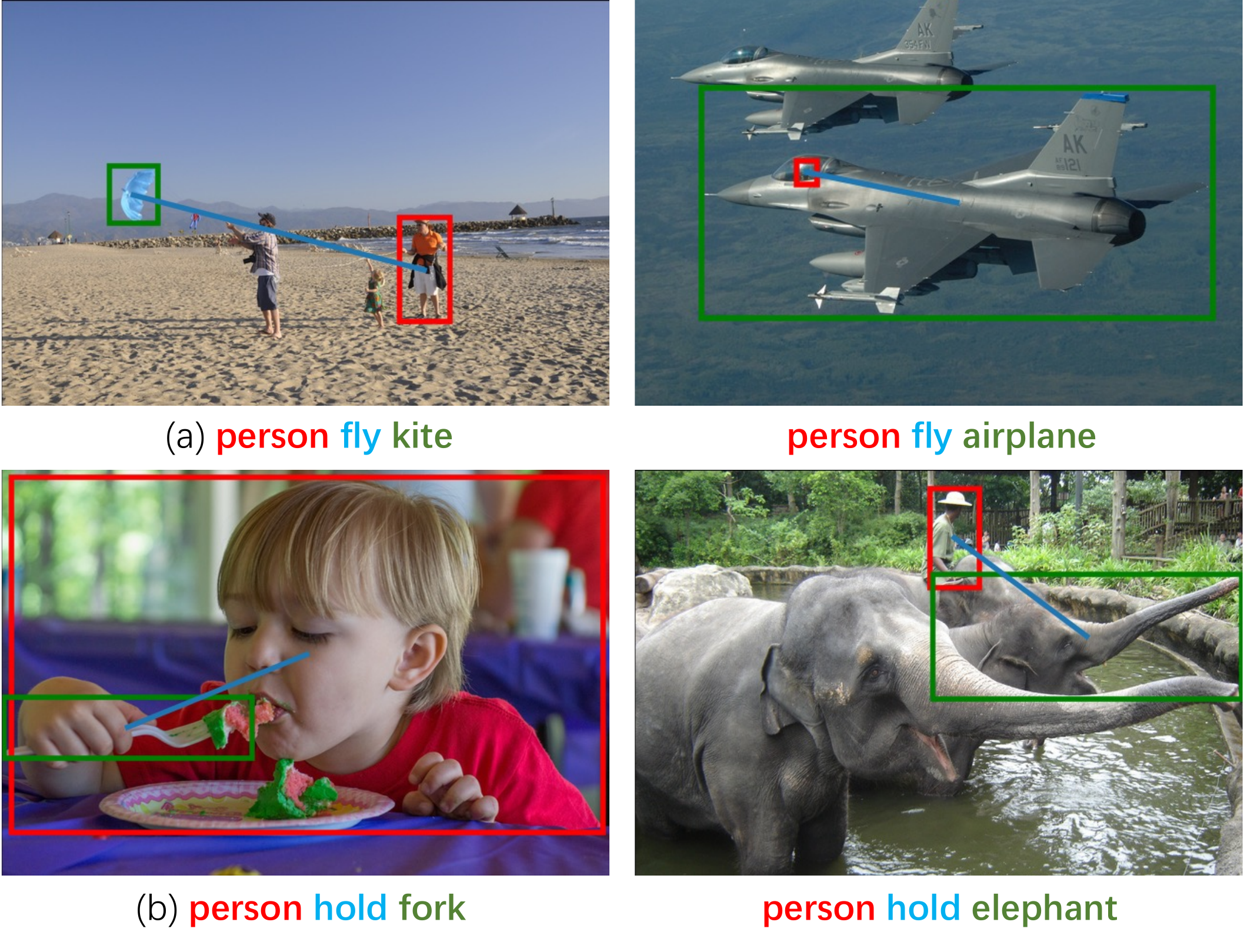}
    \caption{
    % Example images from HICO-DET dataset~\cite{chao2018learning}.
    Interaction classification is inherently challenging.
    In (a), ``fly'' is semantically polysemic, resulting in different objects, poses and relative positions.
    In (b), when ``hold'' is associated with different objects, the appearance, scene background, and human poses are largely different.
    % In (c), even when ``repair'' is associated with the same ``clock'', there still exists large variance in the appearance of human, object and the scene.
}
\vspace{-8pt}
\label{fig:intro-cases}
\end{figure}

Human-Object Interaction (HOI) detection has attracted a lot of interests in recent years~\cite{gupta2015visual, chao2018learning, gkioxari2018detecting, gao2018ican, liao2020ppdm, tamura2021qpic}. The task consists of two sub-tasks. The first is human and object detection. It is usually performed by common object detection methods. The second is \emph{interaction classification} of each human-object (HO) pair. This sub-task is very challenging due to the complex appearance variations in the interaction categories. See ~\cref{fig:intro-cases} for examples. It is the focus of most previous HOI methods, as well as this work.

Most previous HOI methods focus on \emph{learning better human-object features}, including modeling relation and context via GNN~\cite{qi2018learning, ulutan2020vsgnet, wang2020contextual, gao2020drg} or attention mechanism~\cite{gao2018ican, ulutan2020vsgnet, zhong2021polysemy}, decoupling localization and classification\cite{zhang2021mining, zhou2022disentangled, liao2022gen}, leveraging vision-language knowledge~\cite{liao2022gen, dong2022catn} and introducing multi-scale feature to transformer~\cite{kim2022mstr}. However, for interaction classification they all adopt the simple linear classifier that performs the dot product of the human-object feature and a \emph{static} weight vector, which represents an interaction category.

In this work, we propose a new approach that enhances the above paradigm and complements most previous HOI methods. It is motivated by the recent work Query2label\cite{liu2021query2label}, a transformer-based classification network. It proposes a new concept we call \emph{category-specific query}. Unlike the queries in other transformer methods, each query is associated to a specific and fixed image category during training and inference. This one-to-one binding makes the query learn to model each category more effectively. The queries are converted to image specific category representations via a transformer decoder. This method achieves excellent performance on multi-label image classification task.

We extend this approach for human-object interaction classification. Essentially, our approach replaces traditional category representation as a static weight vector in previous HOI methods with category queries learnt as described above. The same linear classifier is adopted. Such category queries are more effective, and \emph{adaptive} for different images, giving rise to better modeling of the complex variations in each interaction category. This is the crucial difference between this work and a simple adaption of ~\cite{liu2021query2label} to HOI. Notably, this work is the first to address the category weight representation problem in the HOI community.

Note that our proposed category specific query is different and not related to those queries in other transformer-based HOI methods\cite{tamura2021qpic,zou2021end,kim2021hotr,chen2021reformulating}. Specifically, category queries extract image-level features as the category representation. The queries in other methods are human-object instance-level features and category-agnostic.

Our method is simple, lightweight and general. The overview is in \cref{fig:overview}. It is complementary to any off-the-shelf HOI method that provides human-object features. The modification of both inference and training is small. The incurred additional cost is marginal. 

In experiments, our approach is validated on three representative and strong HOI baseline methods, two transformer-based methods~\cite{tamura2021qpic, liao2022gen} and a traditional two-stage method~\cite{zhang2021spatially}. They are all significantly improved by our approach. New state-of-the-art results are obtained on two benchmarks. Specifically, we obtain 36.03 mAP on HICO-DET. Comprehensive ablation studies and in-depth discussion are also provided to verify the effectiveness of implementation details in our approach. It turns out that our method is more effective on challenging images that contain more human-object instances, a property that is rarely discussed by previous HOI methods.

\section{Related Work}

\subsection{Instance Query Learning in HOI Detection}
DETR\cite{carion2020end} firstly proposes the concept of object instance query for object detection task. Such queries essentially learn the priors of both object appearance and spatial location. DETR leverages those queries to probe image features through a transformer~\cite{vaswani2017attention} and localize unique objects in the image. Motivated by its great success, many works~\cite{tamura2021qpic, zou2021end,kim2021hotr,zhang2021mining,liao2022gen,zhou2022disentangled} adapt such detection transformer framework to HOI detection by simply treating the HOI triplet~\cite{tamura2021qpic, zou2021end} or H-O pair~\cite{kim2021hotr, chen2021reformulating} as an object. A few of them~\cite{qu2022distillation, dong2022catn, zhong2022hardquerymining} pay attention to adapting the plain query to this task. DOQ~\cite{qu2022distillation} proposes a knowledge distillation model using oracle queries to facilitate the representation learning of a transformer-based detector; HQM~\cite{zhong2022hardquerymining} explicitly constructs hard positive queries from ground truth to train the model to be less vulnerable to spacial variations; CATN~\cite{dong2022catn} utilizes the object category prior generated from external object detector and language model for query initialization. In summary, these transformer-based methods use each query to aggregate context information not restricted to one interaction category, in order to predict a potential HOI instance at a specific location.

%In DETR\cite{carion2020end} and its variants\cite{tamura2021qpic}\cite{zou2021end} in HOI,  use object/HOI query to predict object/HOI instance. Each query learns a embedding vector and extracts image features related to its embedding content through a transformer decoder. However, its feature extraction is not stable and efficient since it is category-agnostic and in practice learns priors of difference category varies hugely in visual appearance.

In DETR\cite{carion2020end} and its variants\cite{tamura2021qpic, zou2021end, kim2021hotr, chen2021reformulating, zhang2021mining, liao2022gen} in HOI, as the queries are category-agnostic, their association to object categories are dynamic and unstable during training. This could be problematic. For example, it is well known that the convergence of DETR training is slow. 
In contrast, our proposed query is category-specific. The learning is guided by image-level classification task and stable. Such queries learn category-specific priors and are good representation for interaction categories.

\subsection{Feature Learning in HOI Detection}
\noindent{\textbf{Early methods.}}
Based on two-stage detection framework, early works make many efforts to help feature learning, including employing architectures effective in modeling relation and context like GNN~\cite{qi2018learning, ulutan2020vsgnet, wang2020contextual, gao2020drg} and attention module~\cite{gao2018ican, ulutan2020vsgnet, zhong2021polysemy}, leveraging fine-grained visual features~\cite{li2019transferable, gupta2019nofrills, kim2020detecting, li2020pastanet, wan2019pose} like human pose and introducing language prior~\cite{zhong2021polysemy, bansal2020functional, liu2020amplifying, kim2020detecting, gao2020drg}.

\noindent{\textbf{Transformer-based methods.}}
% The recently proposed DETR~\cite{carion2020end} succeeds in object detection and inspires a lot of works\cite{tamura2021qpic, zou2021end, kim2021hotr, chen2021reformulating, zhong2022hardquerymining, zhang2022STIP, zhang2022upt, iftekhar2022ssrt} on HOI detection. DETR formulates object detection as set prediction and introduce the transformer architecture~\cite{vaswani2017attention} to solve that in an end-to-end manner. It learns a set of object queries to directly model objects in an image, removing the handcrafted anchor design.
Motivated by DETR\cite{zou2021end}, many methods\cite{tamura2021qpic, zou2021end, kim2021hotr, chen2021reformulating} leverage transformer architecture\cite{vaswani2017attention} and extend the object query in DETR to HOI query. With the help of HOI query and transformer's built-in attention mechanism, those methods learn effective feature representation for HOI triplet or H-O pair.

Based on those pioneer transformer-based methods, recently, many methods are proposed to further help feature learning, by decoupling H-O pair localization and interaction classification~\cite{zhang2021mining, zhou2022disentangled, liao2022gen} or exploiting multi-scale feature in transformer architecture~\cite{kim2022mstr}. Some works~\cite{liao2022gen, dong2022catn} leverage vision-language knowledge in CLIP~\cite{radford2021CLIP} or design a pretrained model~\cite{yuan2022rlip} specifically for HOI; others utilize information like human poses~\cite{wu2022bodypartmap} or spatial configurations~\cite{iftekhar2022ssrt} that has been used in early HOI detectors.

\noindent{\textbf{Relation to the proposed method.}} Previous methods learn the H-O feature while ours learns the category query as the category representation feature. Thus, they are complementary. The interaction classification is simply by measuring the similarity between the two types of features. The integration of our method to previous HOI methods is simple.

\section{Our Method}
\label{sec.method}

\begin{figure*}
  \centering
  \includegraphics[width=0.9\linewidth]{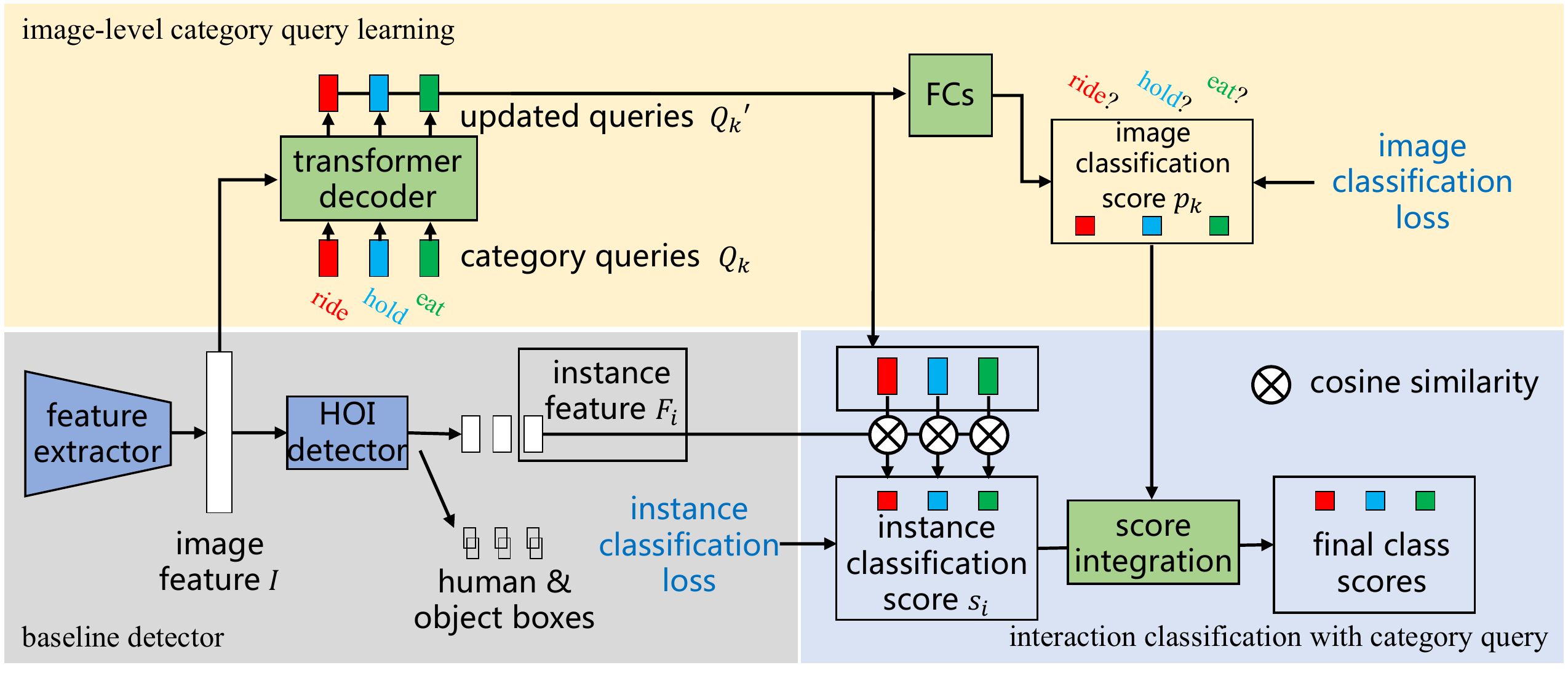}
  \caption{Overview of our method.
  It consists of two components (top and bottom right). It can be integrated with any baseline HOI method (bottom left) that provides image feature $I$ and human-object instance features $F_i$. See \cref{sec.method} for details.
   }
  \vspace{-8pt}
  \label{fig:overview}
\end{figure*}

The overview of our method is in \cref{fig:overview}. It consists of two components, the image-level category query learning (top block) and human-object interaction classification (bottom right block).

The first component is detailed in \cref{sec:category_query_learning}. It is briefly summarized here. A number of queries (embedding vectors) are associated to human-object interaction category, in a one-to-one manner. Such queries interact with image features (provided by a baseline HOI method) through a transformer decoder~\cite{vaswani2017attention} and become image-specific queries. Learning of both the queries and decoder weights is supervised by an auxiliary image-level classification task. In this way, the queries are learnt to capture \textbf{category-specific} feature and become good feature representation for these categories. Besides some minor details, this step is the same as the previous work Query2label~\cite{liu2021query2label}, which is for multi-label image classification task.

%\Cref{fig:attn} shows that the query learns effective category priors. 

The second component is detailed in \cref{sec:interaction_classification_with_category_query}. For the first time, we adopt the category query learning method for human-object interaction classification tasks. The cosine similarity between the category query and human-object feature is used for interaction classification. Thus, it works with any HOI method that provides human-object features. Besides, the image-level classification results turn out moderately helpful in an score integration step, which is an extra technique that benefits the performance.

Overall, the proposed method is simple, effective, lightweight and general. It can be combined with most previous HOI detection methods (bottom left block in \cref{fig:overview}), with small modification, as elaborated in \cref{sec:integration}.

\subsection{Image-level Category Query Learning}
\label{sec:category_query_learning}
Similar as in Query2Label\cite{liu2021query2label}, for $K$ human-object interaction categories, we define their one-to-one corresponding category queries, which are learnable embedding vectors, $\{Q_{k}\} \in \mathbb{R}^{K \times D}$, where $D$ is the vector dimension. 

Each query $Q_{k}$ aggregates image features $\mathbf{I} \in \mathbb{R}^{H \times W \times D}$ through a transformer decoder and is updated to image specific query ${Q'_{k}}$,
\begin{equation}
    \{Q'_{k}\} = \operatorname{decoder}(\{Q_{k}\}, I). 
    \label{eq.feat_extraction}
\end{equation}

Note that the decoder structure has several variants, which are studied in~\cite{cheng2022mask2former}. Our experiments show that the structure is of minor importance, as discussed in \cref{subsec:ablation}. Specifically, our decoder consists of two layers, each of which consisting of a cross-attention layer, a self-attention layer and a FFN layer, in order.

Then, each image-level classification probabilities $p_{k}$ is computed by applying a category-specific fully-connected layer and a sigmoid activation on the updated query $Q'_k$.

\begin{equation}
    p_{k} = \operatorname{sigmoid}(\operatorname{FC}(Q'_{k}))
    \label{eq.img_classification}
\end{equation}

Learning of the category query $\{Q_k\}$ and the decoder weights is supervised by common image classification losses. To deal with the label imbalance problem, focal loss~\cite{lin2017focal} and asymmetric loss (ASL)~\cite{ridnik2021asymmetric} are used. Asymmetric loss is a variant of focal loss. It is more robust for high label imbalance and noises. Our experiments (see \cref{subsec:ablation}) show that it is slightly better.

Specifically, with the classification probability $p_{k}$ and the shifted probability $p'_{k} = \max \left(p_{k} - m, 0 \right)$, the asymmetric loss is
\begin{equation}
    \mathcal{L}_{img} = 
    \frac{1}{K} \sum_{k=1}^{K}
    \left\{\begin{array}{ll}
    \left( 1-p_{k} \right)^{\gamma+} \log \left( p_{k} \right), & y_{k}=1, \\
    \left( p'_{k} \right) ^{\gamma-} \log \left( 1 - p'_{k} \right), & y_{k}=0,
    \end{array}\right.
\label{eq.img_cls_loss}
\end{equation}
where the binary label $y_{k}$ indicates the existence of category $k$ in the image, and $\gamma+$, $\gamma-$ as well as $m$ are hyper-parameters. We use the default values in ASL\cite{ridnik2021asymmetric}, $\gamma+ = 0$, $\gamma- = 4$ and $m = 0.05$. 

In this way, the category queries are learnt to encode the category priors. \Cref{fig:attn} is the visualization of the heatmaps in the cross-attention layer of the decoder. Each category query learns to locate the human body parts related to discriminative feature of its corresponding interaction category, e.g., in \cref{fig:attn-c}, the query of ``hold'' highlights the hand region, while in the same image the query of ``ride'' highlights the foot region. It qualitatively demonstrates that the query learning is effective in encoding \textbf{category-specific} information.

The updated queries adaptively extract category-related features for each image, with the help of the transformer decoder's built-in multi-head cross-attention layer.

\subsection{Interaction Classification with Category Query}
\label{sec:interaction_classification_with_category_query}

In this step, we apply the updated category queries $\{Q'_k\}$ as the weights for interaction classification. Given the $i$-th human-object instance, its classification probability score for $k$-th interaction category is simply the cosine similarity between its feature $F_{i}$ and the category query feature $Q'_{k}$

\begin{equation}
   s_{i,k} = \operatorname{sigmoid}(\frac{Q'_{k} \cdot F_{i}}{|Q'_{k}| \times |F_{i}|}).
   \label{eq.dynamic_w}
\end{equation}

There are no restrictions or assumptions on the human-object feature $F_{i}$. Most previous HOI methods should be applicable here.

In this step, the traditional classification weight from static parameters is replaced with the category query adaptive on each image. This behavior is essentially different from ~\cite{liu2021query2label}, which uses category queries as image feature.

\begin{figure}
  \centering
  \begin{subfigure}{\linewidth}
    \includegraphics[width=\linewidth]{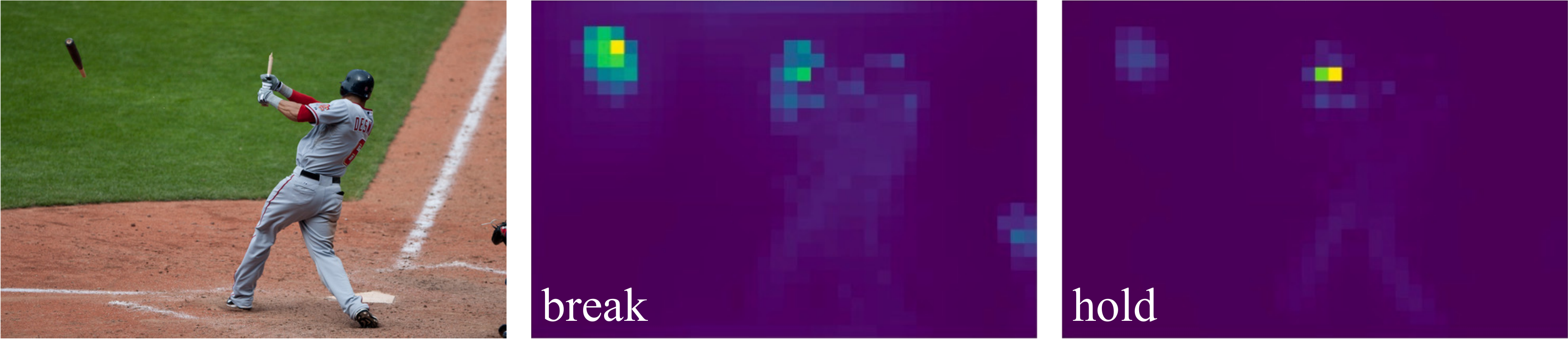}
    \caption{Image with ``person-break-baseball bat'' and ``hold-baseball bat'.}
    \label{fig:attn-a}
  \end{subfigure}
  \hfill
  \begin{subfigure}{\linewidth}
    \includegraphics[width=\linewidth]{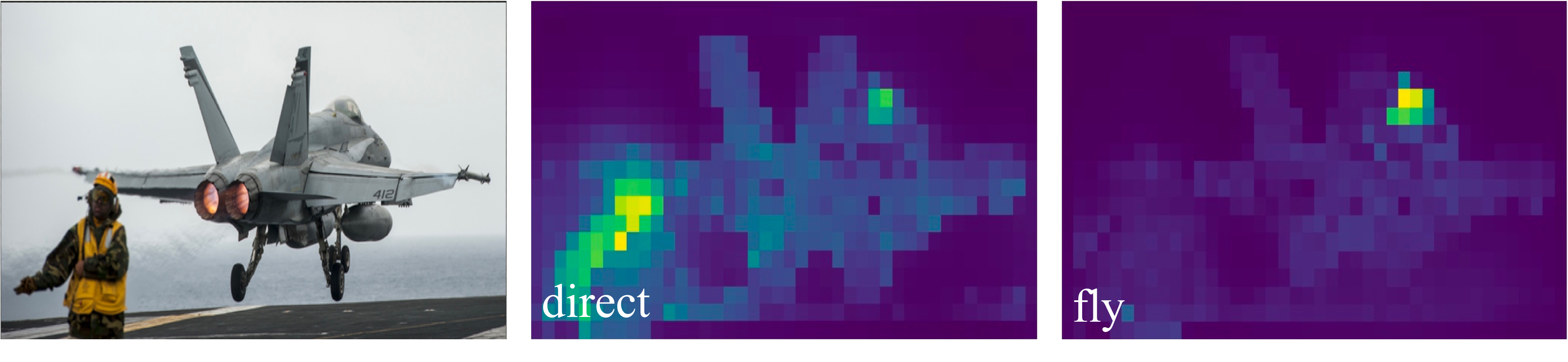}
    \caption{Image with ``person-direct-airplane'' and ``fly-airplane''.}
    \label{fig:attn-b}
  \end{subfigure}
  \hfill
  \begin{subfigure}{\linewidth}
    \includegraphics[width=\linewidth]{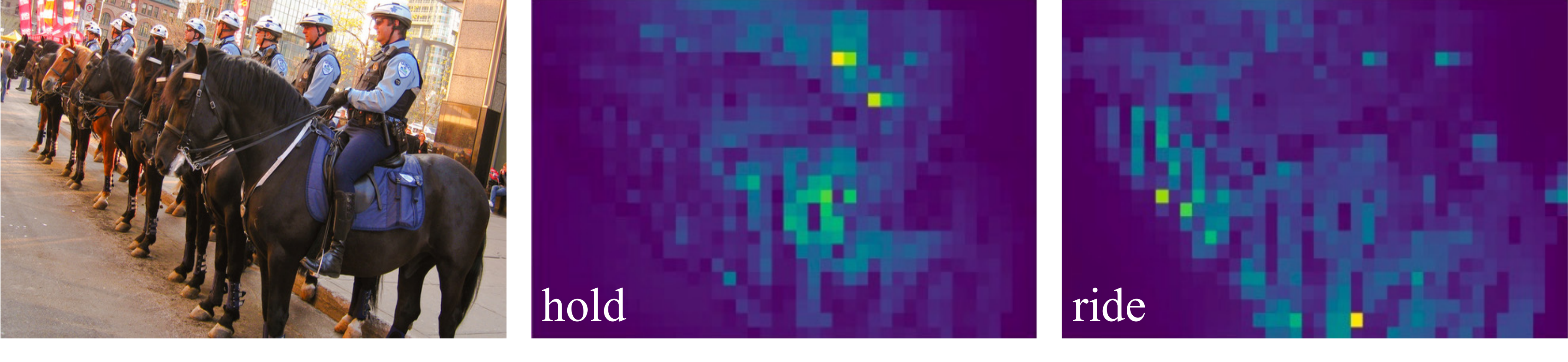}
    \caption{Image with ``person-hold-horse'' and ``ride-horse''.}
    \label{fig:attn-c}
  \end{subfigure}
  \caption{From left to right: image, attention maps of the cross-attention layer in the decoder for different interaction categories.}
  \vspace{-8pt}
  \label{fig:attn}
\end{figure}

By ``adaptive'', we mean that the queries are updated dynamically according to the image contents.
As exemplified in \cref{fig:attn-a} and \cref{fig:attn-c}, the queries of ``hold'' learn to highlight different interactive areas in different images and update themselves with features from these areas via attention. This shows the query learning is \textbf{adaptive to image}.

As discussed in \cref{subsec:ablation}, the classification step in ~\cref{eq.dynamic_w} is crucial to make the category query learning effective on human-object interaction classification. Without this step, the image-level category query learning using only image-level classification from Query2label~\cite{liu2021query2label} is of little use.

\noindent{\textbf{Score integration step.}}
The segmentation method in~\cite{he2022rankseg} discards certain categories during pixel classification that have low image classification scores. Motivated by this method, we take a similar score integration step. The image-level classification score $\{p_{i}\}$ is used to enhance the human-object instance classification. The idea is that, the instance score should be higher if the image-level probability is higher. Our implementation is similar as in~\cite{he2022rankseg}. During both training and testing, for each image, the top-$\kappa$ categories ($\kappa = 70$ in this work) with higher image classification scores $\{p_{i}\}$ are selected. The instance score $s_{i,k}$ is slightly modulated such that it becomes higher if the rank of category $k$ is higher. This strategy gives rise to moderate improvement, as verified in \cref{tab:ablation-components}. We left the implementation details in the \supp.

\begin{table*}
  \centering
  \begin{adjustbox}{width=\linewidth}
  \begin{tabular}{c c c | c c c | c c | c c}
    \hline
    % \toprule
    & & & \multicolumn{3}{c |}{HICO-DET} & \multicolumn{2}{c |}{V-COCO} & \multicolumn{2}{c}{Efficiency} \\
    Method & Pipeline & E2E & Full & Rare & Non-Rare & S1 & S2 & \#Params & FPS \\
    \hline
    % \midrule
    \rowg
    QPIC~\cite{tamura2021qpic} & transformer & \cmark & 28.93 & 21.62 & 31.12 & 61.39 & 63.65 & 41M & 19.5 \\
    + \textit{Ours} & transformer & \cmark & 31.08\color{red}(+2.15) & 23.90 & 33.22 & 63.67\color{red}(+2.28) & 65.49 & 46M\color{blue}(+5M) & 18.3\color{blue}(-6.2\%) \\
    \rowg
    SCG~\cite{zhang2021spatially} & two-stage & \xmark & 31.28 & 24.16 & 33.40 & 56.93 & 62.51 & 57M & 4.5 \\
    + \textit{Ours} & two-stage & \xmark & 32.74\color{red}(+1.46) & 26.25 & 34.68 & 59.14\color{red}(+2.21) & 65.61 & 64M\color{blue}(+7M) & 4.1\color{blue}(-8.9\%) \\
    \rowg
    GEN-VLKT~\cite{liao2022gen} & transformer & \xmark & 33.69 & 29.94 & 34.81 & 64.89 & 66.74 & 42M & 21.7 \\
    + \textit{Ours} & transformer & \xmark & 35.36\color{red}(+1.67) & 32.97 & 36.07 & 66.40\color{red}(+1.51) & 69.17 & 47M\color{blue}(+5M) & 20.6\color{blue}(-5.1\%) \\
    \hline
    % \bottomrule
  \end{tabular}
  \end{adjustbox}
  \caption{The performance numbers of three different baseline HOI methods with and without integration of our method, on two datasets. ``E2E'' denotes whether a HOI detector is end-to-end. All models are tested on Tesla V100.}
  \vspace{-8pt}
  \label{tab:exp-improvement}
\end{table*}

\section{Integration to Off-the-shelf HOI Detectors}
\label{sec:integration}

As shown in \cref{fig:overview}, our method is ready to integrate with any baseline HOI method that provides image feature $\mathbf{I}$ and human-object instance feature $\{F_{i}\}$. The integration is simple. During inference, the human-object instance interaction classification part is replaced by our method in \cref{sec:interaction_classification_with_category_query}, the top and bottom right block in \cref{fig:overview}.

During training, the original loss in the baseline HOI method $\mathcal{L}_{base}$ is added to our image classification loss in \cref{eq.img_cls_loss}. The final loss $\mathcal{L}$ for training is

\begin{equation}
  \mathcal{L} = \mathcal{L}_{base} + \lambda * \mathcal{L}_{img},
  \label{eq.image_loss}
\end{equation}
where the weight $\lambda$ is $1.0$ by default. All other hyper parameters and details during training remain the same as in the baseline HOI method.

Thus, our method is general and applicable to most existing HOI methods. In this work, we select three representative yet different baseline methods to verify the effectiveness of our approach, as described below.

\noindent{\textbf{QPIC~\cite{tamura2021qpic}}} is the first to introduce transformer method into HOI task. It is also the baseline for many recent works~\cite{zhang2021mining, zhou2022disentangled, park2022consistency, iftekhar2022ssrt, zhong2022hardquerymining}. Its performance is much better than early one-stage~\cite{liao2020ppdm, kim2020uniondet, wang2020learning} and two-stage~\cite{gupta2019nofrills, gao2020drg, kim2020detecting} methods while keeping a simple and end-to-end architecture. It consists of a CNN backbone as well as a transformer encoder and decoder.

During our integration, the feature map in its transformer encoder is used as the image feature $\mathbf{I}$. The human-object feature $\{F_{i}\}$ is the query feature in its decoder.

\noindent{\textbf{SCG~\cite{zhang2021spatially}}} is a traditional two-stage method and the best in this category. It is also one of the best method that does not use transformer. It uses a multi-stream graph neural network(GNN) for interaction classification. In our experiment, the detection boxes are from a fine-tuned detector provided by DRG~\cite{gao2020drg} for HICO-DET and a fine-tuned DETR for V-COCO.

During our integration, the CNN feature map in the backbone of SCG is used as image feature $\mathbf{I}$. The human-object feature $\{F_{i}\}$ is generated through RoI pooling with detected human and object boxes and fused with the GNN.

\noindent{\textbf{GEN-VLKT~\cite{liao2022gen}}} is also transformer-based, but not end-to-end as pairwise NMS~\cite{zhang2021mining} is used for post-processing.
% Another difference with QPIC is that it utilize the vision-language knowledge in the large-scale pretrained CLIP~\cite{radford2021CLIP}, which significantly boosts its performance.
It is the current state-of-the-art method. It uses two parallel decoders for object detection and interaction classification, namely instance decoder and interaction decoder.
% Note that it is not end to end.

During our integration, the feature map in its transformer encoder is used as image feature $\mathbf{I}$. The query feature in the interaction decoder is used as the human-object feature $\{F_{i}\}$. Note that, unlike the majority of HOI detection methods, the original GEN-VLKT uses HOI categories rather than interaction categories during interaction classification. Our experiments still use interaction categories, in order to be consistent with most other methods.

\begin{table*}
  \centering
  \begin{adjustbox}{width=0.8\linewidth}
  \begin{tabular}{c | c c | c c c | c c c}
    \toprule
     & & & \multicolumn{3}{c |}{Default} & \multicolumn{3}{c}{Known Object} \\
    Method & Detector & Backbone & Full & Rare & Non-rare & Full & Rare & Non-rare \\
    \midrule
    % Two-stage methods: \\
    % GPNN~\cite{qi2018learning} & COCO & ResDCN152 & 13.11 & 9.34 & 14.23 & - & - & - \\
    % IP-Net~\cite{wang2020learning} & COCO & Hourglass104 & 19.56 & 12.79 & 21.58 & 22.05 & 15.77 & 23.92 \\
    % FCMNet~\cite{liu2020amplifying} & COCO & ResNet50 & 20.41 & 17.34 & 21.56 & 22.04 & 18.97 & 23.12 \\
    % ACP~\cite{kim2020detecting} & COCO & ResNet152 & 20.59 & 15.92 & 21.98 & - & - & - \\
    % PD-Net\cite{zhong2021polysemy} & COCO & ResNet152-FPN & 20.81 & 15.90 & 22.28 & 24.78 & 18.88 & 26.54 \\
    DRG\cite{gao2020drg}	& HICO-DET & ResNet50-FPN & 24.53 & 19.47 & 26.04 & 27.98 & 23.11 & 29.43 \\
    % IDN~\cite{li2020hoianalysis} & HICO-DET & ResNet50 & 26.29 & 22.61 & 27.39 & 28.24 & 24.47 & 29.37 \\
    % SCG\cite{zhang2021spatially} (with DRG det) & HICO-DET & ResNet50-FPN & 31.33 & 24.72 & 33.31 & 34.37 & 27.18 & 36.52 \\
    % \midrule
    % One-stage methods: \\
    % UnionDet\cite{kim2020uniondet} & COCO & ResNet50-FPN & 17.58 & 11.72 & 19.33 & 19.76 & 14.68 & 21.27 \\
    % PPDM\cite{liao2020ppdm} & HICO-DET & Hourglass104 & 21.94 & 13.97 & 24.81 & 24.81 & 17.09 & 27.12 \\
    % HOI-Trans\cite{zou2021end} & HICO-DET & ResNet50 & 23.46 & 16.91 & 25.41 & 26.15 & 19.24 & 28.22 \\
    GG-Net\cite{zhong2021glance} & HICO-DET & Hourglass104 & 23.47 & 16.48 & 25.60 & 27.36 & 20.23 & 29.48 \\
    % VCL~\cite{hou2020visual} & HICO-DET & ResNet50 & 23.63 & 17.21 & 25.55 & 25.98 & 19.12 & 28.03 \\
    % DRG\cite{gao2020drg}	& HICO-DET & ResNet50-FPN & 24.53 & 19.47 & 26.04 & 27.98 & 23.11 & 29.43 \\
    % HOTR~\cite{kim2021hotr} & HICO-DET & ResNet50 & 25.10 & 17.34 & 27.42 & - & - & - \\
    IDN~\cite{li2020hoianalysis} & HICO-DET & ResNet50 & 26.29 & 22.61 & 27.39 & 28.24 & 24.47 & 29.37 \\
    % AS-Net\cite{chen2021reformulating} & HICO-DET & ResNet50 & 28.87 & 24.25 & 30.25 & 31.74 & 27.07 & 33.14 \\
    QPIC\cite{tamura2021qpic} & HICO-DET & ResNet50 & 29.07 & 21.85 & 31.23 & 31.68 & 24.14 & 33.93 \\
    % QPIC\cite{tamura2021qpic} & HICO-DET & ResNet101 & 29.90 & 23.92 & 31.69 & 32.38 & 26.06 & 34.27 \\
    % CPC~\cite{park2022consistency} & HICO-DET & ResNet50 & 29.63 & 23.14 & 31.57 & - & - & - \\
    SCG~\cite{zhang2021spatially} & HICO-DET & ResNet50-FPN & 31.33 & 24.72 & 33.31 & 34.37 & 27.18 & 36.52 \\
    CDN\cite{zhang2021mining} & HICO-DET & ResNet50 & 31.78 & 27.55 & 33.05 & 34.53 & 29.73 & 35.96 \\
    % CDN\cite{zhang2021mining} & HICO-DET & ResNet101 & 32.07 & 27.19 & 33.53 & 34.79 & 29.48 & 36.38 \\
    % UPT~\cite{zhang2022upt} & HICO-DET & ResNet50 & 31.66 & 25.94 & 33.36 & 35.05 & 29.27 & 36.77 \\
    % UPT~\cite{zhang2022upt} & HICO-DET & ResNet101 & 32.31 & 28.55 & 33.44 & 35.65 & 31.60 & 36.86 \\
    DT~\cite{zhou2022disentangled} & HICO-DET & ResNet50 & 31.75 & 27.45 & 33.03 & 34.50 & 30.13 & 35.81 \\
    STIP~\cite{zhang2022STIP} &  HICO-DET & ResNet50 & 31.60 & 27.75 & 32.75 & 34.41 & 30.12 & 35.69 \\
    % CATN~\cite{dong2022catn} & HICO-DET & ResNet50 & 31.86 & 25.15 & 33.84 & 34.44 & 27.69 & 36.45 \\
    HQM~\cite{zhong2022hardquerymining} & HICO-DET & ResNet50 & 32.47 & 28.15 & 33.76 & - & - & - \\
    MSTR~\cite{kim2022mstr} & HICO-DET & ResNet50 & 31.17 & 25.31 & 32.92 & 34.02 & 28.83 & 35.57 \\
    RLIP~\cite{yuan2022rlip} & COCO+VG & ResNet50 & 32.84 & 26.85 & 34.63 & - & - & - \\
    IF~\cite{liu2022interactiveness} & HICO-DET & ResNet50 & 33.51 & 30.30 & 34.46 & 36.28 & 33.16 & 37.21 \\
    GEN-VLKT-B~\cite{liao2022gen} & HICO-DET & ResNet50 & 33.75 & 29.25 & 35.10 & 36.78 & 32.75 & 37.99 \\
    GEN-VLKT-M~\cite{liao2022gen} & HICO-DET & ResNet101 & 34.78 & 31.50 & 35.77 & 38.07 & 34.94 & 39.01 \\
    GEN-VLKT-L~\cite{liao2022gen} & HICO-DET & ResNet101 & 34.95 & 31.18 & \textbf{36.08} & \textbf{38.22} & 34.36 & \textbf{39.37} \\
    BodyPartMap~\cite{wu2022bodypartmap} & HICO-DET & ResNet50 & \textbf{35.15} & \textbf{33.71} & 35.58 & 37.56 & \textbf{35.87} & 38.06 \\
    \midrule
    GEN-VLKT-B + \textit{Ours} & HICO-DET & ResNet50 & 35.36 & 32.97 & 36.07 & 38.43 & 34.85 & 39.50 \\
    GEN-VLKT-M + \textit{Ours} & HICO-DET & ResNet101 & 35.83 & 32.91 & 36.70 & 38.79 & 35.28 & \textbf{39.84} \\
    GEN-VLKT-L + \textit{Ours} & HICO-DET & ResNet101 & \textbf{36.03} & \textbf{33.16} & \textbf{36.89} & \textbf{38.82} & \textbf{35.51} & 39.81 \\
    \bottomrule
  \end{tabular}
  \end{adjustbox}
  \caption{
  The proposed method achieves state-of-the-art on HICO-DET~\cite{chao2018learning}. The best results are marked in \textbf{bold}.
  }
  \vspace{-8pt}
  \label{tab:exp-sota-hicodet}
\end{table*}

\begin{table}
  \centering
  \begin{adjustbox}{width=\linewidth}
  \begin{tabular}{c c | c c}
    \toprule
    Method & Backbone & Scenario \#1 & Scenario \#2 \\
    \midrule
    % Two-stage Method: \\
    % InteractNet~\cite{gkioxari2018detecting} & 40.0 & - \\
    % GPNN~\cite{qi2018learning} & 44.0 & - \\
    % iCAN~\cite{gao2018ican} & 45.3 & 52.4 \\
    % UnionDet~\cite{kim2020uniondet} & 47.5 & 56.2 \\
    % TIN~\cite{li2019transferable} & 47.8 & 54.2 \\
    % VCL~\cite{hou2020visual} & 48.3 & - \\
    DRG~\cite{gao2020drg} & R50FPN & 51.0 & - \\
    % IP-Net~\cite{wang2020learning} & 51.0 & - \\
    % VSGNet\cite{ulutan2020vsgnet} & 51.8 & 57.0 \\
    % PMFNet~\cite{wan2019pose} & 52.0 & - \\
    % PD-Net~\cite{zhong2021polysemy} & 52.6 & - \\
    % CHGNet~\cite{wang2020contextual} & 52.7 & - \\
    % IDN~\cite{li2020hoianalysis} & 53.3 & 60.3 \\
    % \midrule
    % One-stage Method: \\
    % UnionDet~\cite{kim2020uniondet} & 47.5 & 56.2 \\
    % PD-Net~\cite{zhong2021polysemy} & 52.6 & - \\
    % HOI-Trans~\cite{zou2021end} & R50 & 52.9 & - \\
    % FCMNet~\cite{liu2020amplifying} & 53.1 & - \\
    % ACP~\cite{kim2020detecting} & R152 & 53.2 & - \\
    % IDN~\cite{li2020hoianalysis} & 53.3 & 60.3 \\
    % AS-Net~\cite{chen2021reformulating} & R50 & 53.9 & - \\
    SCG~\cite{zhang2021spatially} & R50 & 54.2 & 60.9 \\
    GG-Net~\cite{zhong2021glance} & HG104 & 54.7 & - \\
    % HOTR~\cite{kim2021hotr} & R50 & 55.2 & 64.4 \\
    QPIC~\cite{tamura2021qpic} & R50 & 58.8 & 61.0 \\
    % QPIC~\cite{tamura2021qpic} & R101 & 58.3 & 60.7 \\
    % CPC~\cite{park2022consistency} & R50 & 63.1 & 65.4 \\
    % UPT~\cite{zhang2022upt} & R50 & 59.0 & 64.5 \\
    % CATN~\cite{dong2022catn} & R50 & 60.1 & - \\
    HQM~\cite{zhong2022hardquerymining} & R50 & 63.6 & - \\
    CDN~\cite{zhang2021mining} & R50 & 61.7 & 63.8 \\
    % CDN-S~\cite{zhang2021mining} & R50 & 61.7 & 63.8 \\
    % CDN-B~\cite{zhang2021mining} & R50 & 62.3 & 64.4 \\
    % CDN-L~\cite{zhang2021mining} & R101 & 63.9 & 65.9 \\
    GEN-VLKT-B~\cite{liao2022gen} & R50 & 62.4 & 64.5 \\
    GEN-VLKT-M~\cite{liao2022gen} & R101 & 63.3 & 65.6 \\
    GEN-VLKT-L~\cite{liao2022gen} & R101 & 63.6 & 65.9 \\
    MSTR~\cite{kim2022mstr} & R50 & 62.0 & 65.2 \\
    BodyPartMap~\cite{wu2022bodypartmap} & R50 & 63.0 & 65.1 \\
    IF~\cite{liu2022interactiveness} & R50 & 63.0 & 65.2 \\
    DT~\cite{zhou2022disentangled} & R50 & \textbf{66.2} & 68.5 \\
    STIP~\cite{zhang2022STIP} & R50 & 65.1 & \textbf{69.7} \\
    \midrule
    % QPIC-R50 + Ours & 64.04 & 66.06 \\
    % QPIC-R101 + Ours & 64.97 & 66.77 \\
    % CDN-B + Ours & 63.32 & 65.31 \\
    % CDN-L + Ours & \textbf{65.01} & \textbf{66.85} \\
    % Our baseline & & \\
    GEN-VLKT-B + \textit{Ours} & R50 & 66.4 & 69.2 \\
    GEN-VLKT-M + \textit{Ours} & R101 & \textbf{66.8} & 69.8 \\
    GEN-VLKT-L + \textit{Ours} & R101 & 66.5 & \textbf{69.9} \\
    \bottomrule
  \end{tabular}
  \end{adjustbox}
  \caption{
  Comparison with state-of-the-art methods on V-COCO~\cite{gupta2015visual} dataset.
  The best results are marked in \textbf{bold}.
  % R-50 and R-101 indicate ResNet50 and ResNet101 backbones.
  }
  \vspace{-8pt}
  \label{tab:exp-sota-vcoco}
\end{table}

\section{Experiments}

In this section, we verify the applicability and effectiveness of the proposed method through experiments. In \cref{subsec:exp-setup}, we introduce the experimental settings. Then we demonstrate the effectiveness of the proposed method over 3 baselines in \cref{subsec:compare-baseline}, and show it achieves SOTA results on major benchmarks in \cref{subsec:compare-sota}. Next, in \cref{subsec:ablation} we conduct comprehensive ablation studies on the key components as well as detailed technical designs. Lastly, we provide some analysis and visualization in \cref{subsec:analysis}.

\subsection{Datasets}
\label{subsec:exp-setup}

HICO-DET\cite{chao2018learning} and V-COCO\cite{gupta2015visual} are two widely-used HOI benchmarks. HICO-DET contains 47,776 images, with 38,118 for training and 9,658 for testing. There are 600 HOI categories in HICO-DET, consisting of 117 interaction classes and 80 object classes. Each HOI category is composed of an interaction and an object. V-COCO is a subset of MS-COCO\cite{lin2014microsoft} with HOI annotations, including 10,346 images (2,533 for training, 2,867 for validation and 4,946 for testing). It has 80 object categories same with HICO-DET and 29 interaction categories.

\noindent{\textbf{Evaluation metrics.}}
For HICO-DET, we adopt the commonly used mAP metric\cite{chao2018learning}. Each prediction is a \textlangle{}human, interaction, object\textrangle{} triplet. A prediction is a true positive only when the human and object bounding boxes both have IoU \textgreater 0.5 w.r.t. ground truth and the interaction classification result is correct. We evaluate the performance in two different settings following\cite{chao2018learning}. In the \textit{known object} setting, for each HOI category, we evaluate the prediction only on the images containing the target object category. In \textit{default} setting, the detection result of each category is evaluated on the full test set. In each setting, we report the mAP over (1) all 600 HOI categoryies (Full), (2) 138 categories with less than 10 training samples (Rare), and (3) the remaining 462 categories (Non-rare). For V-COCO, we use the role mAP following\cite{gupta2015visual}, under both scenario \#1 (including objects) and \#2 (ignoring objects). The performance is evaluated using its official evaluation toolkit.

\subsection{Improvement on Three Different Baselines}
\label{subsec:compare-baseline}
\cref{tab:exp-improvement} summarizes the performance of the three baseline HOI methods before and after integration of our method. Backbone is ResNet50. All these methods are significantly improved. Specifically, QPIC~\cite{tamura2021qpic} is improved by 2.15 mAP, making it competitive with those more recent works~\cite{kim2022mstr, zhang2021mining, zhang2022upt}. SCG~\cite{zhang2021spatially} is improved by 1.46 mAP, demonstrating that our method is not limited to transformer-based baselines. The current SOTA method GEN-VLKT~\cite{liao2022gen} is improved by 1.67 mAP, producing the new SOTA result (also refer to \cref{tab:exp-sota-hicodet}).

On the V-COCO dataset~\cite{gupta2015visual}, the performance improvement is similar, which is 2.28, 2.21 and 1.51 mAP on QPIC, SCG and GEN-VLKT, respectively.

Notably, for SCG~\cite{zhang2021spatially}, as the object detector is fixed during the training of its interaction classification network, the improvement by our method is purely due to better interaction classification, not a better fine-tuned CNN backbone or a better object detector. This further consolidates that the category query learning is effective.

To verify that the performance improvement is not due to a larger model, we also compare the model size and running speed. Our method increases the model parameters by a few millions, which is small compared to the original model size. The running speed, measured by FPS, is only decreased by a few percent. The marginal additional cost shows that our method is quite lightweight.

\subsection{Comparison with State-of-the-art}
\label{subsec:compare-sota}
\cref{tab:exp-sota-hicodet} and \cref{tab:exp-sota-vcoco} compare our method with many previous methods for HICO-DET and V-COCO datasets, respectively. GEN-VLKT~\cite{liao2022gen} is used as our baseline.

On HICO-DET, our result with ResNet50 backbone already outperforms all previous methods under both \textit{default} and \textit{known object} settings. With the stronger ResNet101 backbone, our method achieves the new state-of-the-art 36.03 full mAP under \textit{default} settings and 38.82 under \textit{known object} settings.

On V-COCO dataset, our method achieves the new state-of-the-art performance on Scenario 1, with an AP of 66.4 for ResNet50, surpassing ~\cite{zhou2022disentangled}. For scenario 2, it is comparable with the state-of-the-art~\cite{zhang2022STIP}.

\begin{table}
  \centering
  \begin{adjustbox}{width=1.0\linewidth}
  \begin{tabular}{c c c c c c c}
    \hline
    % \toprule
    & C1 & C2 & C3 & Full & Rare & Non-Rare \\
    \hline
    % \midrule
    \rowg
    a & - & - & - & 33.69 & 29.94 & 34.81 \\
    b & \cmark & - & - & 33.86 \color{red}(+0.17) & 31.12 & 34.68 \\
    \rowg
    c & \cmark & \cmark & - & 34.98 \color{red}(+1.29) & 31.73 & 35.95 \\
    d & \cmark & \cmark & \cmark & \textbf{35.36} \color{red}(+1.67) & \textbf{32.97} & \textbf{36.07} \\
    \hline
    % \bottomrule
  \end{tabular}
  \end{adjustbox}
  \caption{Ablation study of several variants of our method, starting from the baseline (a) to our approach (d).
  \textbf{C1}, \textbf{C2} and \textbf{C3} are described in \cref{subsec:ablation}.
  The best results are marked in \textbf{bold}.
  }
  \vspace{-8pt}
  \label{tab:ablation-components}
\end{table}

\begin{table}
  \centering
  % \begin{adjustbox}{width=0.8\linewidth}
  \begin{tabular}{c c | c c c}
    \hline
    % \toprule
    loss type & $\lambda$ & Full & Rare & Non-Rare \\
    \hline
    % \midrule
    - & 0 & \cellg 34.21 & \cellg 30.15 & \cellg 35.42 \\
    \hline
    % \midrule
    focal loss~\cite{lin2017focal} & 0.5 & 34.43 & 31.06 & 35.44 \\
    & 1.0 & \cellg 34.51 & \cellg 31.08 & \cellg 35.53 \\
    & 1.5 & 34.35 & 31.54 & 35.19 \\
    & 2.0 & \cellg 34.29 & \cellg 31.18 & \cellg 35.22 \\
    \hline
    % \midrule
    % simplified ASL~\cite{liu2021query2label} & 0.5 & & \\
    % & 1.0 & & \\
    % & 1.5 & & \\
    % & 2.0 & & \\
    % \midrule
    ASL~\cite{ridnik2021asymmetric} & 0.5 & 34.57 & 31.91 & 35.36 \\
    & 1.0 & \cellg \textbf{34.98} & \cellg 31.73 & \cellg \textbf{35.95} \\
    & 1.5 & 34.77 & \textbf{32.08} & 35.57 \\
    & 2.0 & \cellg 34.41 & \cellg 31.92 & \cellg 35.15 \\
    \hline
    % \bottomrule
  \end{tabular}
  % \end{adjustbox}
  \caption{Ablation on the type and weight $\lambda$ of the image classification loss.
  The best results are marked in \textbf{bold}.
  }
  \vspace{-8pt}
  \label{tab:ablation-imgloss}
\end{table}

\begin{table}
  \centering
  % \begin{adjustbox}{width=0.7\linewidth}
  \begin{tabular}{c | c c c}
    \hline
    % \toprule
    Layer structure & Full & Rare & Non-Rare \\
    \hline
    % \midrule
    \rowg
    S$\rightarrow$C$\rightarrow$F & 34.73 & \textbf{32.09} & 35.52 \\
    C$\rightarrow$S$\rightarrow$F & \textbf{34.98} & 31.73 & \textbf{35.95} \\
    \rowg
    C$\rightarrow$F & 34.63 & 31.34 & 35.61 \\
    \hline
    % \bottomrule
  \end{tabular}
  % \end{adjustbox}
  \caption{Ablation on structure of each layer in the category decoder. Here ``S'', ``C'' and ``F'' stands for the self-attention, cross-attention and FFN in a standard transformer~\cite{vaswani2017attention}.
  \vspace{-8pt}
  }
  \label{tab:ablation-decoderstructure}
\end{table}

\begin{table}
  \centering
  % \begin{adjustbox}{width=0.5\linewidth}
  \begin{tabular}{c | c c c}
    \hline
    % \toprule
    $L$ & Full & Rare & Non-Rare \\
    \hline
    % \midrule
    \rowg
    1 & 34.66 & 31.47 & 35.61 \\
    2 & \textbf{34.98} & \textbf{31.73} & \textbf{35.95} \\
    \rowg
    3 & 34.86 & 31.56 & 35.85 \\
    \hline
    % \bottomrule
  \end{tabular}
  % \end{adjustbox}  
  \caption{Ablation on the number of layers in the category decoder.
  }
  \vspace{-8pt}
  \label{tab:ablation-numlayer}
\end{table}

\subsection{Ablation Experiments}
\label{subsec:ablation}

We perform various ablation experiments to validate the effectiveness of different components in our method. HICO-DET dataset and GEN-VLKT~\cite{liao2022gen} baseline are used.

The proposed method can be divided into 3 components: \textbf{C1} means applying Query2Label~\cite{liu2021query2label} to the baseline detector as a multi-task learning (with feature extractor shared). In detail, it adds the queries, the decoder and image classification loss (\cref{eq.feat_extraction}, \cref{eq.img_classification} and \cref{eq.image_loss}). \textbf{C2} means using the learned query in \textbf{C1} as adaptive interaction classification weight (\cref{eq.dynamic_w}). This is the key component of the proposed method, which makes a distinction between the proposed method and a simple adaption of Query2label~\cite{liu2021query2label} to HOI task. \textbf{C3} denotes the score integration step in \cref{sec:interaction_classification_with_category_query}.

To validate our approach in ~\cref{fig:overview}, several variants with these components in our approach are experimented and summarized in \cref{tab:ablation-components}.
First, (b) is a simple combination of Query2label~\cite{liu2021query2label} and the baseline detector (a) in a multi-task setting. (b) is only slightly better than (a), showing that simply applying Query2Label to HOI is barely helpful.

Second, variant (c) significantly boosts (b), indicating using the queries as adaptive classification weights is the key to the performance improvement. This shows the effectiveness of our most crucial technical design in \textbf{C2}: applying the learned category queries as adaptive interaction classification weights.

Finally, the complete approach is (d). It further adds the integration step of image classification and instance classification score on (c). This technique produces moderate improvement over (c), i.e., 1.67 mAP vs. 1.29 mAP.

\noindent{\textbf{Image classification loss.}}
\cref{tab:ablation-imgloss} compares different loss functions and weights.
First, when $\lambda = 0$, which means no image-level supervision is applied, the improvement over the baseline (33.69 mAP) drops to only 0.52 mAP. This demonstrates the image classification supervision is essential to make the category query learning effective, with either focal loss or ASL.
Additionally, ASL is slightly better than focal loss. By default, $\lambda = 1.0$ is adopted.

\noindent{\textbf{Is asymmtric loss the key?}}
In the ablation above, we can see that ASL does help our image-level query learning. However, it is not the major reason for the performance improvement. To figure out this, we replace focal loss in the plain baseline GEN-VLKT with ASL and the result is only slightly better by 0.08 mAP.

\noindent{\textbf{Decoder structure.}}
\cref{tab:ablation-decoderstructure} compares several structures of the decoder. Compared to the standard  decoder~\cite{vaswani2017attention, carion2020end} (S$\rightarrow$C$\rightarrow$F in the table), putting cross-attention first (C$\rightarrow$S$\rightarrow$F) is slightly better (by 0.25 mAP) without extra computation.
If we remove self-attention (C$\rightarrow$F), the performance drops by 0.35 mAP compared with the S$\rightarrow$C$\rightarrow$F setting. This is probably because self-attention helps to learn the dependencies between different category queries.

\cref{tab:ablation-numlayer} compares different numbers of decoder layers, denoted as $L$. We find that $L=2$ is sufficient. More layers do not help the performance.

\begin{figure}
  \centering
  \includegraphics[width=\linewidth]{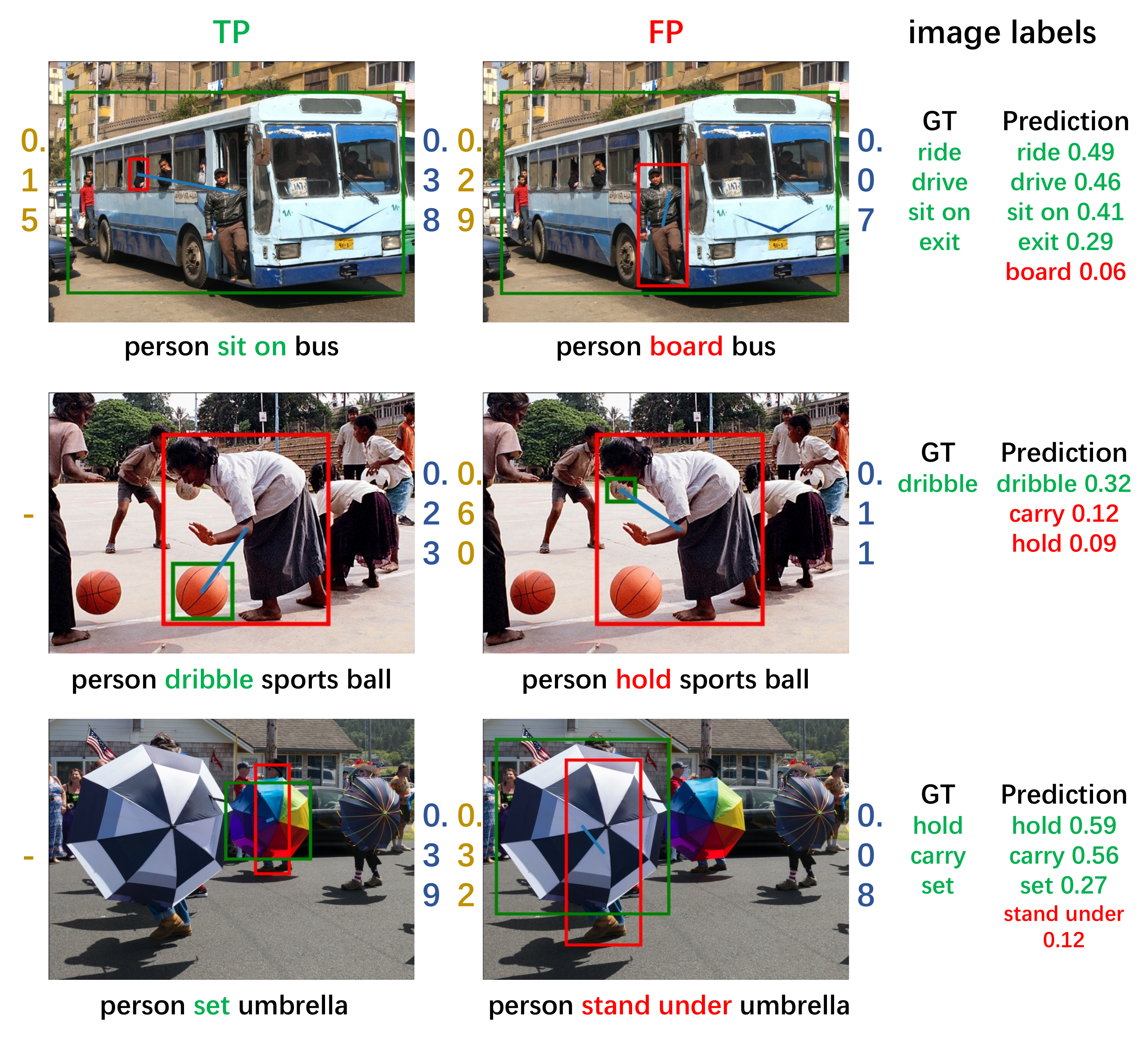}
  \caption{Some qualitative comparison between the baseline and the proposed method on HICO-DET.
  From left to right, column 1: true positive (\textcolor{Green}{TP}) detection results, whose interaction score is increased by our method; column 2: false positive (\textcolor{red}{FP}) detection results, whose interaction score is decreased by our method; column 3: corresponding image-level GT and predictions by our method. Scores on the left and right of an image are the interaction classification scores of the visualized instance from \textcolor{olive}{the baseline} and \textcolor{Violet}{our method}. Best viewed in color. More in \supp.
  }
  \vspace{-8pt}
  \label{fig:qualitative}
\end{figure}

\begin{figure}
  \centering
  \begin{subfigure}{0.9\linewidth}
    % \fbox{\rule{0pt}{2in} \rule{.9\linewidth}{0pt}}
    \includegraphics[width=\linewidth]{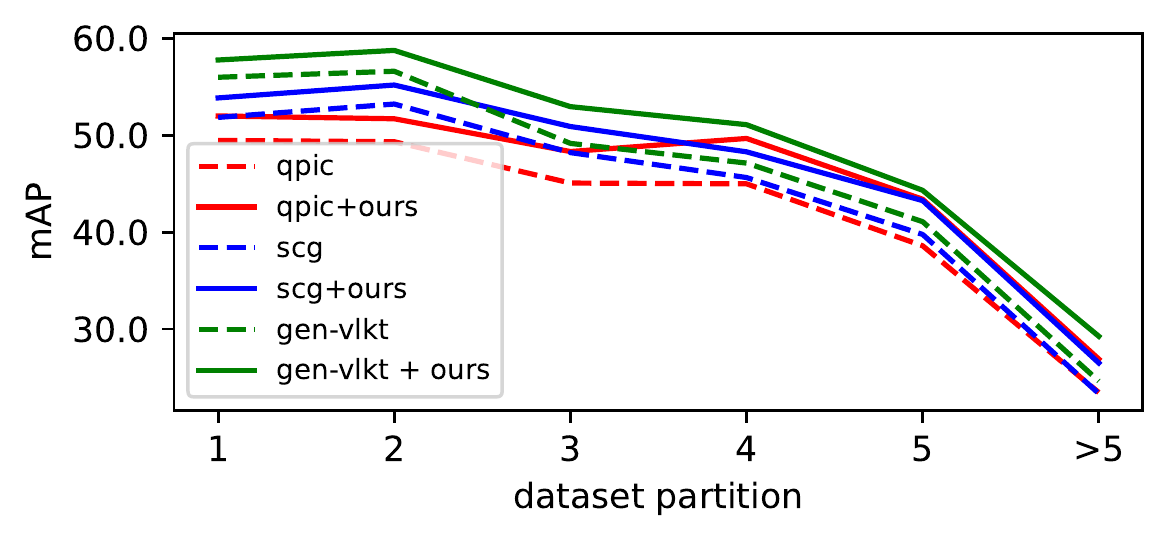}
    \vspace{-4mm}
    % \caption{mAP of the baselines and improved models over different $n$.}
    \label{fig:analysis-a}
  \end{subfigure}
  \begin{subfigure}{0.9\linewidth}
    % \fbox{\rule{0pt}{2in} \rule{.9\linewidth}{0pt}}
    \includegraphics[width=\linewidth]{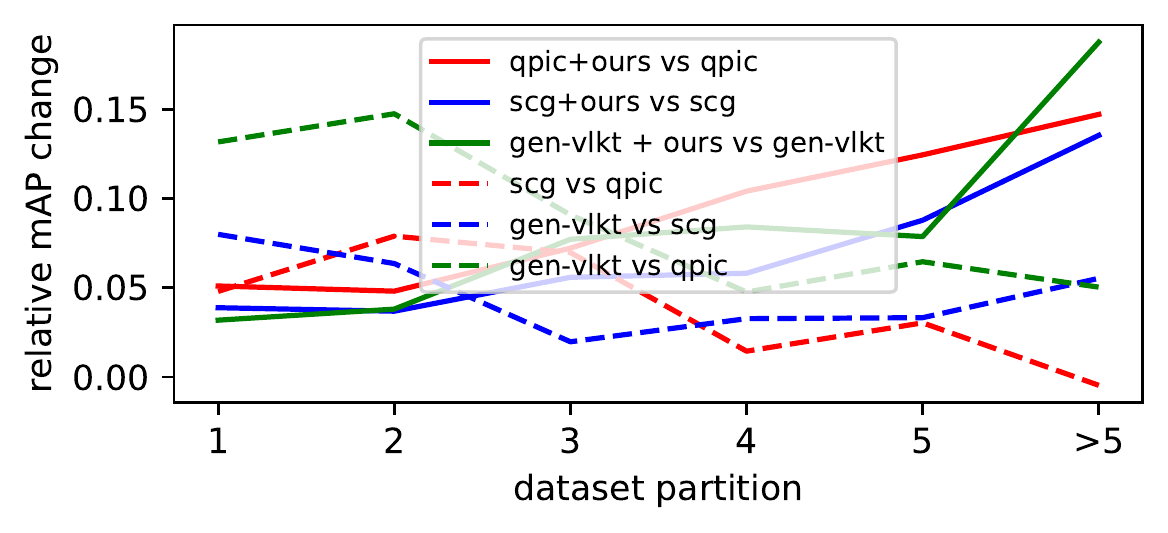}
    % \caption{Relative change of mAP between models over different $n$.}
    \label{fig:analysis-c}
  \end{subfigure}
  \vspace{-8mm}
  \caption{Performance evaluated on image partitions with different $n=1,2,3,4,5,>5$ instances for an interaction category. Top: mAP numbers for three baselines (dashed) as well as our integration (solid). Bottom: relative mAP ratios for improvement produced by our method (solid) and between two arbitrary previous methods (dashed).
  }
  \vspace{-8pt}
  \label{fig:densityanalysis}
\end{figure}

\subsection{Discussions and Analysis}
\label{subsec:analysis}
To understand why the category query learning is effective for human-object interaction classification, we provide some analysis and qualitative results. 

The attention maps in the cross-attention layer of the decoder is visualized in \cref{fig:attn}. For different category queries, the corresponding attention maps show they learn to capture the semantics of the category, while being adaptive to different images. For example, in \cref{fig:attn-a}, the broken part of the bat in the air is highlighted for ``break'' while the left part in the hand is highlighted for ``hold''. This is similar for \cref{fig:attn-b}. In \cref{fig:attn-c}, the attention map highlights many instances with corresponding action ``hold'' and ``ride''.

To qualitatively demonstrate how our method helps, we visualize some cases of the baseline and the proposed method in \cref{fig:qualitative}. In the first case, the baseline predicts a TP of ``person sit on bus'' with a low score 0.15, and a FP of ``person board bus'' with a high score 0.29. Our model also predicts the same TP and FP, but lowers the FP score to 0.07 and lifts the TP score to 0.38. Besides, our model correctly predicts image-level scores: the wrong category ``board'' is given a low score of 0.06 while the four correct categories are given high scores.
In the second case, our model successfully predicts the TP ``person dribble'' missed by the baseline and suppress the FP ``person hold sports'' from 0.6 to 0.11 with the help of correct image-level classification result. This is similar for the third case.

Last, as our category query learning is performed on image level, we conjecture that it is more helpful for images with dense human-object interactions. In such images, an human-object instance is relatively small and hard to learn good feature on its own. However, it may benefit more from the global image level category query feature, which aggregates more information from other similar instances in this image. To validate this conjecture, we partition the images according to their ``interaction density'' and check whether our method produces larger improvement on images that are ``denser''. Specifically, for each interaction category, its mAP is evaluated separately on six different image partition subsets, where each image contains different numbers (n=1, 2, 3, 4, 5 and $>5$) of human-object instances of this category. The mAP results of the three baseline methods and their integrated versions (as in ~\cref{tab:exp-improvement}) are shown in \cref{fig:densityanalysis} (top, dashed vs. solid lines). It shows that: 1) the mAP is lower for larger n, indicating the ``denser'' images are more challenging; 2) our method improves the baselines consistently on all different partitions.

To check whether the proposed method is more effective on ``denser'' images, we use the relative mAP improvement, which is a ratio, $\frac{mAP_{ours}-mAP_{baseline}}{mAP_{baseline}}$, for analysis. The ratio curves of the three baselines are shown in \cref{fig:densityanalysis} (bottom, solid lines). It is clear that the relative improvement becomes larger for larger $n$. This indicates that the image-level category query learning is more effective on these challenging dense images.

To further verify that this behavior is not commonly true, we also compute the relative ratio of three more comparisons, ``gen-vlkt vs. qpic'', ``gen-vlkt vs. scg'' and ``scg vs. qpic'', in which the former outperforms the latter. These curves are also shown in \cref{fig:densityanalysis} (bottom, dashed lines). There is no clear pattern in these curves, indicating that the performance gap between two arbitrary HOI methods are in general not related to the image ``density''.

\section{Conclusion}
This work proposes a novel approach for the human-object interaction classification sub-task in HOI detection. We study the problem of interaction category modeling, in contrast to most previous methods focusing on human-object feature learning. We adopt the concept of category query in a previous method~\cite{liu2021query2label} for HOI, for the first time, and show that it is simple, general and highly effective.

Clearly, this idea of category query modeling is not limited to multi-label image classification and HOI detection. We hope it is useful for other vision tasks.

\noindent{\textbf{Acknowledgment.}}
This work was supported in part by the National Natural Science Foundation of China under Grant 62076183, 61936014 and 61976159, in part by the Natural Science Foundation of Shanghai under Grant 20ZR1473500, in part by the Shanghai Science and Technology Innovation Action Project of under Grant 20511100700 and 22511105300, in part by the Shanghai Municipal Science and Technology Major Project under Grant 2021SHZDZX0100, and in part by the Fundamental Research Funds for the Central Universities. The authors would also like to thank the anonymous reviewers for their careful work and valuable suggestions.

%%%%%%%%% REFERENCES
{\small
\bibliographystyle{ieee_abbrev}
\bibliography{ms}
}

\end{document}

% \title{Image-Adaptive Human Object Interaction Classification}
\title{Supplementary Material for \\
Category Query Learning for Human-Object Interaction Classification}

\author{
Chi~Xie$^1$ \quad
Fangao Zeng$^{2}$ \quad
Yue~Hu$^3$ \quad
Shuang~Liang$^{1}$ \quad
Yichen~Wei$^{2}$ \\
$^{1}${Tongji University} \quad
$^{2}${MEGVII Technology} \quad
$^{3}${Shanghai Jiao Tong University} \\
$^{1}${\tt\small \{chixie, shuangliang\}@tongji.edu.cn} \\ \quad
$^{2}${\tt\small zfg472988436@163.com, wei\_yi\_chen@hotmail.com} \quad
$^{3}${\tt\small 18671129361@sjtu.edu.cn} \\
}

% \maketitle

%%%%%%%%% BODY TEXT

\section{Overview}

In this supplemental file, we provide more details of our work to supply the main paper.
Specifically,
\vspace{2pt}
\begin{itemize}
    \item[$\blacktriangleright$] \textbf{Score integration technique} used in our paper are explained in \cref{sec:score-integration};
    \item[$\blacktriangleright$] \textbf{Implementation details} are summarized in \cref{sec:implementation};
    % \item[$\blacktriangleright$] \textbf{Additional ablations} are presented in \cref{sec:additional-ablation}, including the component ablation on the other 2 baselines and the score integration technique;
    \item[$\blacktriangleright$] \textbf{Additional ablations} are presented in \cref{sec:additional-ablation}, which includes the ablations on the score integration technique;
    \item[$\blacktriangleright$] \textbf{Additional qualitative results} are presented in \cref{sec:additional-qualitative}.
\end{itemize}

\section{Score Integration Technique}
\label{sec:score-integration}

We introduce the score integration step briefly in Sec. 3.2 of the main paper, which leverages the image-level classification scores to stress or suppress certain categories during instance-level interaction categories.
As the score integration step is not the major contribution of the proposed method, and brings minor improvement (as in Tab. 4 of the paper), we did not elaborate on its details in the paper.

Before applying this score integration step, based on Eq. (4) in the paper, we can compute the classification scores for the $i$-th human-object instance over $K$ interaction categories as
\begin{equation}
    s_{i} = \operatorname{sigmoid}\left(
    \left[ \frac{\left(F_{i}, \overline{Q'}_{1}\right)}{\left\| F_{i} \right\| \left\| \overline{Q'}_{1} \right\| },
    \cdots,
    \frac{\left(F_{i}, \overline{Q'}_{K}\right)}{\left\| F_{i} \right\| \left\| \overline{Q'}_{K} \right\| } \right]
    \right),
\end{equation}
where the sigmoid operation is applied on the vector element-wise.

Next, we provide the detailed design of this score integration step. It includes a hard integration and a soft one.

\subsection{Hard Score Integration}
This hard score integration is motivated by the rank-adaptive pixel classification in RankSeg\cite{he2022rankseg}.
It consists of two steps: the first is to use the image classification results to sort and select some interaction categories, and perform H-O pair classification only on the selected categories, namely, category selection; the second is to adopt a series of temperature parameters that ranks the interaction classification results of sorted and selected categories, namely, category ranking.

\noindent{\textbf{Category selection.}} 
Instead of choosing the labels for an H-O pair from all $K$ predefined categories, based on the previous multi-label image classification prediction $\{p_{k}\}$ for the image, we perform a selected-label classification.
First, the top $\kappa$ of the classification weights $\{Q'_{k}\}$ is selected according to the descending order of image classification predictions as
\begin{equation}
    \left[\overline{Q'}_{1}, \cdots, \overline{Q'}_{\kappa}\right]
    = \operatorname{Top}-\kappa\left(\left[Q'_{1}, \cdots, Q'_{K}\right], \{p_{k}\}\right),
\end{equation}
and H-O pair classification is performed as
\begin{equation}
    s_{i}^{h} = 
    \operatorname{sigmoid}\left(
    \left[ \frac{\left(F_{i}, \overline{Q'}_{1}\right)}{\left\| F_{i} \right\| \left\| \overline{Q'}_{1} \right\| }, \cdots, \frac{\left(F_{i}, \overline{Q'}_{\kappa}\right)}{\left\| F_{i} \right\| \left\| \overline{Q'}_{\kappa} \right\| } \right]
    \right),
\label{eq.category_selection}
\end{equation}
where $\left[\overline{Q'}_{1}, \cdots, \overline{Q'}_{\kappa}\right]$ denotes the top $\kappa$ selected category queries (classification weights) associated with the largest $\kappa$ image classification scores, $s_{i}^{h}$ denotes the classification scores with hard score integration, and $\kappa$ represents the number of selected category queries, chosen as a much smaller value than $K$.

\noindent{\textbf{Category ranking.}}
On top of category selection, we apply a set of learnable temperature parameters $[\tau_{1}, \tau_{2}, \cdots, \tau_{\kappa}]$ to adjust the classification scores over the selected top $\kappa$ categories, so ~\cref{eq.category_selection} is changed to
\begin{equation}
    s_{i}^{h} = 
    \operatorname{sigmoid}\left(
    \left[ \frac{\left(F_{i}, \overline{Q'}_{1}\right)}{\left\| F_{i} \right\| \left\| \overline{Q'}_{1} \right\| \tau_{i} }, \cdots, \frac{\left(F_{i}, \overline{Q'}_{\kappa}\right)}{\left\| F_{i} \right\| \left\| \overline{Q'}_{\kappa} \right\| \tau_{\kappa} } \right]
    \right).
\end{equation}

We analyze the influence of $\kappa$ choices and the benefits of such a ranking adjustment in the ablation study.
Note that this is similar to the rank-adaptive pixel classification performed in RankSeg~\cite{he2022rankseg} for image and video segmentation tasks, though their classification is a single-label problem and softmax is applied while ours are multi-label and sigmoid is used.

\subsection{Soft Score Integration}

Another way to utilize the image-level classification scores is to directly multiply the instance classification scores $s_i$ with the image classification probabilities $\{p_{k}\}$, as
\begin{equation}
    s_{i}^{s} = \left[
    \sqrt{s_{i,1} * p_{1}}, \cdots, \sqrt{s_{i,K} * p_{K}}
    \right],
\end{equation}
where $s_{i}^{s}$ denotes the interaction classification scores of the $i$-th H-O instance, with soft sore integration.

Compared with the hard score integration, no interaction class is deprecated during instance classification. They are just stressed or suppressed in a soft way. Therefore, we call this soft score integration.

Note that hard and soft score integration can be applied together, as
\begin{equation}
    s_{i}^{s,h} = \left[
    \sqrt{s_{i,1}^{h} * \overline{p}_{1}}, \cdots, \sqrt{s_{i,\kappa}^{h} * \overline{p}_{\kappa}}
    \right],
\end{equation}
where $\left[ \overline{p}_{1}, \cdots, \overline{p}_{\kappa} \right]$ is the top $\kappa$ in $\{p_{k}\}$.
Through experiments in \cref{tab:ablation-rankingtechnique}, we find both soft and hard integration bring a small improvement and the best result is achieved when both is used.

\section{Implementation Details}
\label{sec:implementation}

Most of the implementation details have been provided in the paper. Here we summarize these details.
In the proposed category query learning, transformer decoder with 2 layers is used by default. The structure of each decoder layer in the proposed decoder consists of a cross-attention module, a self-attention module and a FFN in order. The weights of the existing losses in the baselines are not changed, and an image loss with loss weight $\lambda = 1.0$ is added to the final loss. For the asymmetric loss in image classification, we adopt $\gamma+ = 0$, $\gamma- = 4$ and $m = 0.05$.
Both hard and soft score integration are used. For category selection and ranking in hard score integration, we set $\kappa$ as 70 for HICO-DET~\cite{chao2018learning}.
Hyper-parameters like learning rate, weight decay, batch size and input image size follow the baseline settings by default.

Following the baseline detectors, the feature extractor is frozen for SCG~\cite{zhang2021spatially}, and updated for QPIC~\cite{tamura2021qpic} and GEN-VLKT~\cite{liao2022gen}.
For the experiments on GEN-VLKT, we change its classification classes from 600 HOI categories to 117 interaction categories for HICO-DET and from 263 to 29 for V-COCO, following most HOI detection methods.
For the experiments on SCG, the detection boxes are from a fine-tuned detector provided by DRG~\cite{gao2020drg} for HICO-DET and a fine-tuned DETR for V-COCO~\cite{gupta2015visual}.
The experiment is conducted on 8 Tesla V100 GPUs.

\section{Additional Ablations}
\label{sec:additional-ablation}

In this part, we perform some additional studies on technical details.

\begin{table}
  \centering
  % \begin{adjustbox}{width=1.0\linewidth}
  \begin{tabular}{c c | c | c c c}
    \toprule
    \multicolumn{2}{c |}{hard integration} & soft integration & \multicolumn{3}{c}{Default} \\
    selection & ranking & & Full & Rare & Non-Rare \\
    \midrule
    - & - & - & 34.98 & 31.73 & 35.95 \\
    \cmark & - & - & 35.09 & 32.98 & 35.72 \\
    \cmark & \cmark & - & 35.24 & 32.67 & 36.01 \\
    - & - & \cmark & 35.18 & 32.23 & 36.06 \\
    \cmark & \cmark & \cmark & \textbf{35.36} & \textbf{32.97} & \textbf{36.07} \\
    \bottomrule
  \end{tabular}
  % \end{adjustbox}
  \caption{Ablation on the techniques (soft and hard score integration) that we elaborate in \cref{sec:score-integration} to utilize image-level classification scores.
  The best results are marked in \textbf{bold}.
  }
  \label{tab:ablation-rankingtechnique}
\end{table}

\begin{table}
  \centering
  % \begin{adjustbox}{width=0.9\linewidth}
  \begin{tabular}{c | c c c c c c}
    \toprule
    $\kappa$ & - & 30 & 50 & 70 & 90 & 117 \\
    \midrule
    mAP & 34.98 & 35.04 & 35.19 & \textbf{35.24} & 35.08 & 35.03 \\
    \bottomrule
  \end{tabular}
  % \end{adjustbox}
  \caption{Ablation on the number of interaction categories in the hard score integration step, i.e., $\kappa$. The metric for comparison is the full mAP under \textit{default} setting on HICO-DET dataset. ``-'' denotes the hard score integration is not used.
  The best results are marked in \textbf{bold}.
  }
  \label{tab:ablation-topkcategories}
\end{table}

\subsection{Integration of Image-level Classification Scores}
As mentioned in \cref{sec:score-integration}, the score integration process is proposed to utilize the image-classification scores in the proposed method, with two strategies: the \textbf{hard score integration}, consisting of category selection and category ranking, and the \textbf{soft score integration}, which is a score multiplication operation between instance-level and image-level classification scores.
As shown in \cref{tab:ablation-rankingtechnique}, each of them brought a marginal improvements: the model with hard score integration achieves 35.24 mAP while the one with score integration achieves 35.18 mAP. Together, a performance of 35.36 is obtained. We use this two techniques together by default.

\subsection{Different $\kappa$ in Hard Score Integration}
In this part, we study to influence of the number of the selected categories, denoted as $\kappa$, in the hard score integration step. As shown in \cref{tab:ablation-topkcategories}, the selection and ranking on interaction categories works best when $\kappa = 70$. For a smaller $\kappa$, some categories may be filtered by mistake, like when $\kappa = 30$, the performance is only 35.04 mAP, falls behind the optimal setting by 0.15 mAP. When $\kappa = 117$, none of the categories are filtered and only ranking operation is still effective. This results in a little performance drop of 0.16 mAP. We use $\kappa = 70$ by default.

\begin{figure*}
  \centering
  \includegraphics[width=0.8\linewidth]{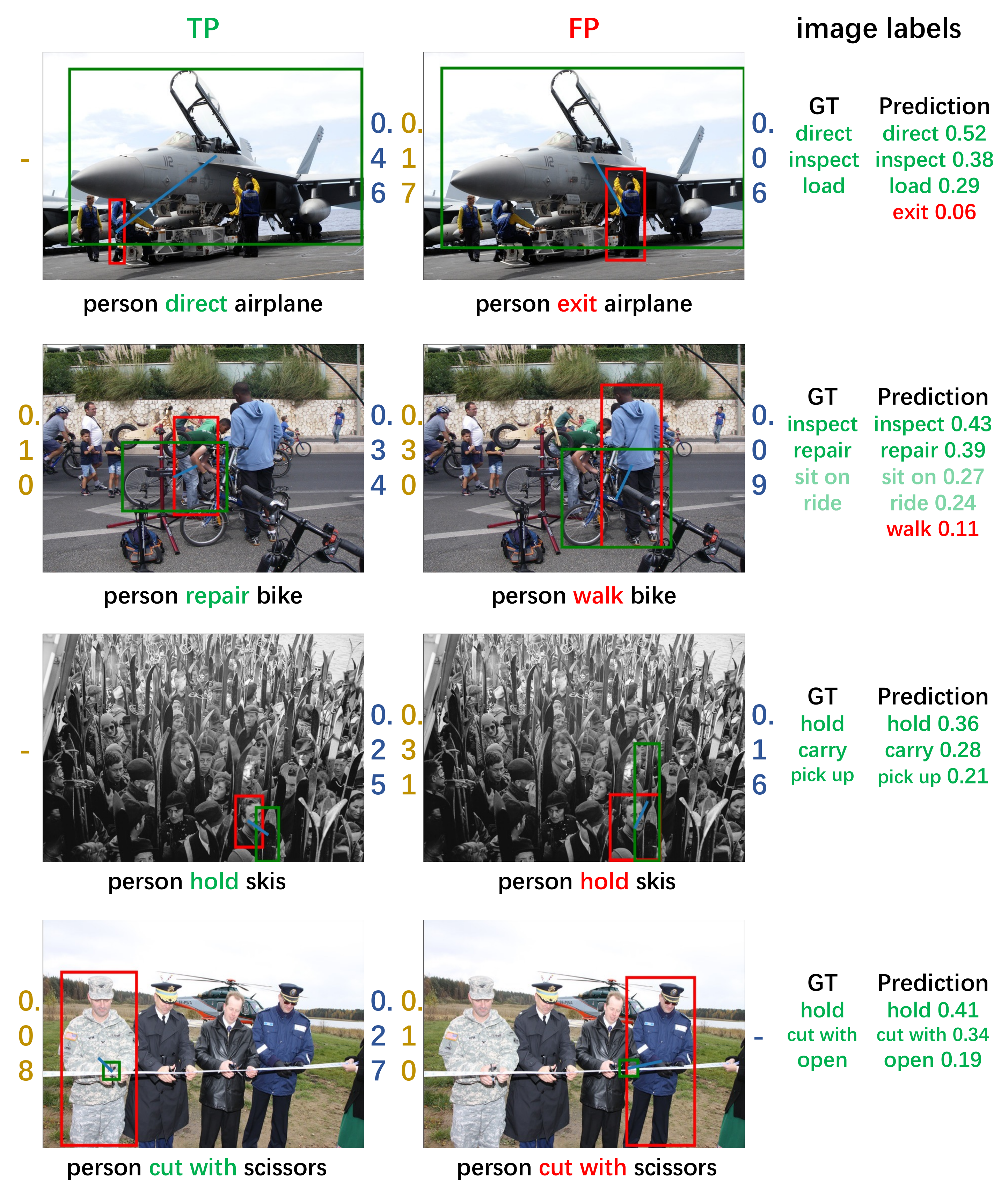}
  \caption{More qualitative comparison between the baseline and the proposed method on HICO-DET.
  From left to right, column 1: true positive (\textcolor{Green}{TP}) detection results, whose interaction score is increased by the proposed method; column 2: false positive (\textcolor{red}{FP}) detection results, whose interaction score is decreased by the proposed method; column 3: corresponding image-level GT and predictions by the proposed method. Scores on the left and right of an image are the interaction classification scores of the visualized instance in the image from \textcolor{olive}{the baseline} and \textcolor{Violet}{the proposed method}. ``-'' for score denotes a instance not discovered (thus no scores).
  Best viewed in color.
  }
  \label{fig:qualitative-more}
\end{figure*}

\section{Additional Qualitative Results}
\label{sec:additional-qualitative}

In \cref{fig:qualitative-more}, we provide more qualitative results in addition to the cases in the paper.

The first case shows the proposed method uses the feature of other instances in the image to help the recognition of a small and challenging instance. The image contains multiple instances of person directing and inspecting an airplane. The TP instance visualized is associated with a small and occluded person, which the baseline fails to discover (the score is denoted as ``-''). The proposed method successfully predict this instance with a high score of 0.46. This is consistent with the quantitative discussion on Fig. 5 in the paper.

In the second case, the proposed method discovers an interaction category ``repair'' neglected by the baseline, possibly with the help of correlations between categories (``inspect'' and ``repair''). The ``repair'' interaction is semantically abstract, but the existence of ``inspect'' may help. This may explain why removing the self-attention from our decoder with cause performance drop in Tab. 6 of the paper: it may learns the dependencies between different interaction categories. In the forth case, the learning of ``cut with'' interaction may also benefit from the recognition of ``hold''. Additionally, there is an obvious annotation mistake in the second case: interactions like ``ride'' and ``sit on'' are not labeled though they exists (in the background). The proposed method still produces relatively high image-level classification scores for these two categories. Actually, such annotation mistakes exists widely in HICO-DET dataset, and the increase on mAP may not fully show the effectiveness of the proposed method.

In the third and forth cases, our method shows its ability to distinguish whether instances belonging to an interaction category existing in the image. In the third image, though it produce a relatively high ``hold'' score of 0.36 at image level, it does not take all the H-O pairs in the image as ``hold'', which would be very wrong. It successfully discovers the TP ``hold'' instance that the baseline missed, and suppresses the FP ``hold'' from 0.31 to 0.16. This is consistent with the quantitative results in Tab.4 of the paper that shows the proposed method benefits more from the \textit{adaptive} instance classification weight rather than simply an image classification task. Notably, these two are challenging images with ``dense'' interaction instances, especially the third case, which corresponds to the discussion in Fig. 5 and Sec 5.5 of the paper.

\section{Potential Limitation and Social Impact}

The proposed method focuses on the interaction classification sub-task in HOI detection. It does not improve H-O pair detection directly. In the future, we will try to extend this idea to the classification of human and objects in HOI to improve H-O pair detection. 

The proposed algorithm has no evident negative impact to society. However, someone might use this method for malicious usage, e.g., to attack people in military usage or invasion of privacy with surveillance. Therefore, we encourage well-intended application of the proposed method.

%%%%%%%%% REFERENCES
{\small
\bibliographystyle{ieee_abbrev}
\bibliography{supplementary}
}